\documentclass[11pt,a4paper]{article}
\usepackage[hyperref]{emnlp-ijcnlp-2019}
\usepackage{times}
\usepackage{latexsym}

\usepackage{url}

\usepackage{graphicx}  %Required
\usepackage{bm}
\usepackage{amsmath}
\usepackage{multirow}
\usepackage{booktabs}

\aclfinalcopy % Uncomment this line for the final submission

\title{A Syntax-aware Multi-task Learning Framework for \\
Chinese Semantic Role Labeling}
\author{Qingrong Xia, Zhenghua Li\thanks{$~$ Corresponding author.}, Min Zhang \\
Institute of Artificial Intelligence, School of Computer Science and Technology, \\ Soochow University, China \\
{\tt kirosummer.nlp@gmail.com,  \{zhli13, minzhang\}@suda.edu.cn }\\
}

\date{}

\begin{document}
\maketitle
\begin{abstract}
Semantic role labeling (SRL) aims to identify the predicate-argument structure of a sentence.
Inspired by the strong correlation between syntax and semantics, previous works pay much attention to improve SRL performance on exploiting syntactic knowledge, achieving significant results.
Pipeline methods based on automatic syntactic trees and multi-task learning (MTL) approaches using standard syntactic trees are two common research orientations.
In this paper, we adopt a simple unified span-based model for both span-based and word-based Chinese SRL as a strong baseline.
Besides, we present a MTL framework that includes the basic SRL module and a dependency parser module.
Different from the commonly used hard parameter sharing strategy in MTL, the main idea is to extract implicit syntactic representations from the dependency parser as external inputs for the basic SRL model.
Experiments on the benchmarks of Chinese Proposition Bank 1.0 and CoNLL-2009 Chinese datasets show that our proposed framework can effectively improve the performance over the strong baselines.
With the external BERT representations, our framework achieves new state-of-the-art 87.54 and 88.5 F1 scores on the two test data of the two benchmarks, respectively.
In-depth analysis are conducted to gain more insights on the proposed framework and the effectiveness of syntax.
\end{abstract}

\section{Introduction}
Semantic role labeling (SRL) is a fundamental and important task in natural language processing (NLP), which aims to identify the semantic structure (\textit{Who} did \textit{what} to \textit{whom}, \textit{when} and \textit{where}, etc.) of each given predicate in a sentence.
Semantic knowledge has been widely exploited in many down-stream NLP tasks, such as information extraction \cite{bastianelli2013textual}, machine translation \cite{liu2010semantic, gao2011corpus} and question answering \cite{shen2007using, wang2015machine}.

\begin{figure}[!t]
	\centering
	\includegraphics[scale=0.8]{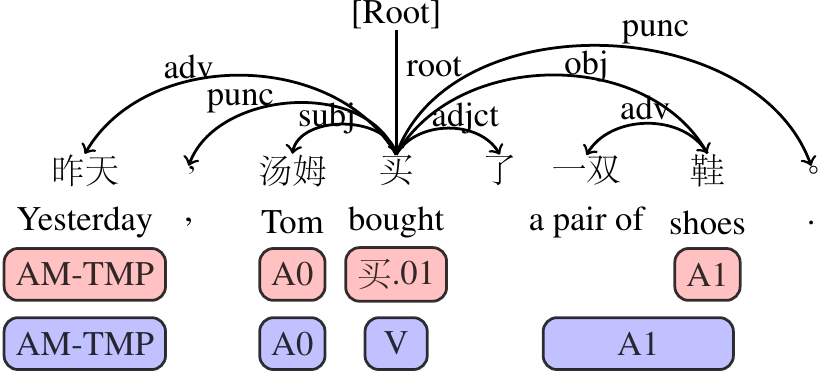}
	\caption{Example of span-based (blue blocks) and word-based (red blocks) SRL formulations in a sentence, where the top part is its dependency tree.} % of SUDA format.}
	\label{fig:srl-example}
\end{figure}

There are two formulations of SRL in the community according to the definition of semantic roles. 
The first is called \textit{span-based} SRL, which employs a continuous word span as a semantic role and follows the manual annotations in the PropBank \cite{palmer2005proposition} and NomBank \cite{meyers2004nombank}.
The second is \textit{word-based} SRL \cite{surdeanu2008conll}, also called \textit{dependency-based} SRL, whose semantic role is usually syntactic or semantic head word of the manually annotated word span. 
Figure \ref{fig:srl-example} gives an example of the two forms in a sentence, where ``bought'' is the given predicate.

Intuitively, syntax and semantics are strongly correlative.
For example, the semantic \textit{A0} and \textit{A1} roles are usually the syntactic subject and object, as shown in Figure \ref{fig:srl-example}.
Inspired by the correlation,  researchers try to improve SRL performance by exploring various ways to integrate syntactic knowledge \cite{roth2016neural, he2018syntax, swayamdipta2018syntactic}.
In contrast, some recent works \cite{he2017deep, tan2017deep, cai2018full} propose deep neural models for SRL without considering any syntactic information, achieving promising results.
Most recently, \citet{he2018jointly, li2019dependency} extend the span-based models to jointly tackle the predicate and argument identification sub-tasks of SRL.

Compared with the large amount of research for English SRL, Chinese SRL works are rare, mainly because of the limited amount of data and lack of attention of Chinese researchers.
For Chinese, the commonly used datasets are Chinese Proposition Bank 1.0 (CPB1.0) (span-based) \cite{xue2008labeling} and CoNLL-2009 Chinese (word-based) \cite{hajivc2009conll}.
The CPB1.0 dataset follows the same annotation guideline with the English PropBank benchmark \cite{palmer2005proposition}.
\citet{wu2015can} present a top model based selection preference approach to improve Chinese SRL.
Since the amount of CPB1.0 dataset is small, \citet{xia2017progressive} exploit heterogeneous SRL data to improve the performance via a progressive learning approach.
The CoNLL-2009 benchmark is released by the CoNLL-2009 shared task \cite{hajivc2009conll}.
Previous works \cite{marcheggiani2017simple, he2018syntax, cai2018full} mainly focus on building more powerful models or exploring the usage of external knowledge on this dataset.

\begin{figure*}[!t]
	\centering
	\includegraphics[scale=0.9]{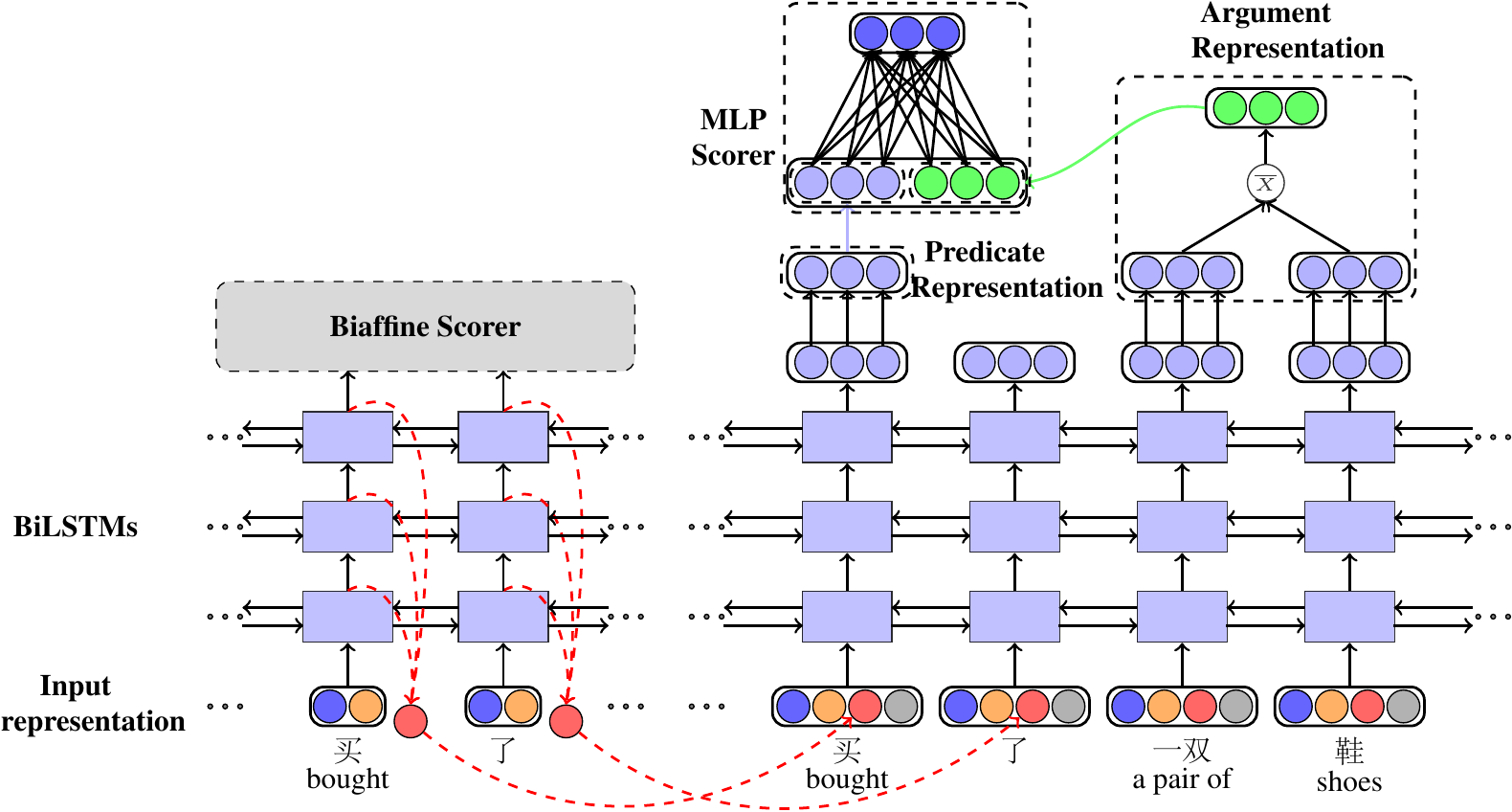}
	\caption{The detailed architecture of our proposed framework, where the left part is the dependency parser and the right part is the basic SRL module, respectively.}
	\label{fig:srl-framework}
\end{figure*}

Inspired by the development of neural models and exploration of syntactic information, this paper proposes a MTL framework to extract syntactic representations as the external input features for the simple unified SRL model.
The contributions of our paper are three-folds:

\begin{enumerate}
\setlength{\topsep}{0pt}
\setlength{\itemsep}{0pt}
\setlength{\parsep}{0pt}
\setlength{\parskip}{0pt}
\setlength{\listparindent}{0em} %段落缩进量
    \item We introduce a simple unified model for span-based and word-based Chinese SRL.
    \item We propose a MTL framework to extract implicit syntactic representations for SRL model, which significantly outperforms the baseline model.
    \item Detailed analysis gains crucial insights on the effectiveness of our proposed framework.
\end{enumerate}

We conduct experiments on the benchmarks of CPB1.0 and CoNLL-2009.
The results show that our framework achieves new state-of-the-art 87.54 and 88.5 F1 scores on the two test data, respectively.

\section{Basic SRL Model}
Motivated by the recently presented span-based models \cite{he2018jointly, li2019dependency} for jointly predicting predicates and arguments, we introduce a simple unified span-based model.
Formally, given a sentence $s = {w_1, w_2, ..., w_n}$, the span-based model aims to predict a set of labeled predicate-argument relationships $\mathcal{Y} \subseteq \mathcal{P} \times \mathcal{A} \times \mathcal{R}$, where $\mathcal{P} = \{w_1, w_2, ..., w_n\}$ is the set of all candidate predicates, $\mathcal{A} = \{(w_i, ..., w_j) | 1 \leq i \leq j \leq n\}$ is the set of all candidate arguments, and $\mathcal{R}$ is the set of the semantic roles.
Following \citet{he2018jointly}, we also include a null label $\epsilon$ in the role set $\mathcal{R}$ indicating no relation between the focused predicate and argument.
The model objective is to optimize the probability of the predicate-argument-role tuples $y \in \mathcal{Y}$ in a sentence $s$, which is formulated as:
\begin{equation}
\begin{split}
P(y|s) &= \prod_{p \in \mathcal{P}, a \in \mathcal{A}, r \in \mathcal{R}} P(y_{(p, a, r)} | s) \\
    &= \prod_{p \in \mathcal{P}, a \in \mathcal{A}, r \in \mathcal{R}} \frac{e^{\phi(p, a, r)}}{\sum_{r' \in \mathcal{R}} e^{\phi(p, a, r')}}
\end{split}
\end{equation}
where $\phi(p, a, r) = \phi_p(p) + \phi_a(a) + \phi_r(p, a)$ is the score of the predicate-argument-relation tuple.
We directly adopt the model architecture of \citet{he2018jointly} as our basic SRL model with a modification on the argument representation.
The architecture of the basic SRL module is shown in the right part of Figure \ref{fig:srl-framework}, and we will describe it in the following subsections.
% As shown in Figure \ref{fig:srl-framework}, the SRL module of our framework contains four parts: 1) input representation for building the token representation $x_i$ for each word $w_i$, 2) highway BiLSTMs are employed to encode the input representations, 3) predicate and argument representations are designed to learn the corresponding representations, and 4) MLP scorers to compute the probability of semantic roles $\mathcal{R}$ of candidate predicates and arguments.

\subsection{Input Layer}
Following \citet{he2018jointly, li2019dependency}, we employ CNNs to encode Chinese characters for each word $w_i$ into its character representation, denoted as $\bm{rep}^{char}_{i}$.
Then, we concatenate $\bm{rep}^{char}_{i}$ with the word embedding $\bm{emb}^{word}_{i}$ to represent the word-level features as our basic model input.
In addition, we also employ BERT representations \cite{devlin2018bert} to boost the performance of our baseline model, which we denote as $\bm{rep}^{BERT}_{i}$.
Formally, the input representation of $w_i$ is:
\begin{equation}
\bm{x}_i = \bm{rep}^{char}_{i} \oplus \bm{emb}^{word}_{i} \oplus \bm{rep}^{BERT}_{i}
\end{equation}
, where $\oplus$ is the concatenation operation.
Our basic SRL model and BERT-enhanced baseline depend on whether including the BERT representation $\bm{rep}^{BERT}_{i}$ or not.
% Moreover, we also encode the syntactic knowledge of the dependency parser module into the input representation, which we will describe later.

\subsection{BiLSTM Encoder}
Over the input layer, we employ the BiLSTMs with highway connections \cite{srivastava2015training, zhang2016highway} to encode long-range dependencies and obtain rich representations， denoted as $\bm{h}_i$ for time stamp $i$.
The highway connections are used to alleviate the gradient vanishing problem when training deep neural networks.

\subsection{Predicate and Argument Representations}
We directly employ the output of the top BiLSTM as the predicate representation at each time stamp.
For all the candidate arguments, we simplify the representations by employing the mean operation over the BiLSTM outputs within the corresponding argument spans, which achieves similar results compared with the attention-based span representations \cite{he2018jointly} on English SRL in our preliminary experiments.
Formally, 
\begin{equation}
\begin{split}
\bm{rep}^{p}_i &= \bm{h}_i \\
\bm{rep}^{a}_{j, k} &= mean(\bm{h}_j, ..., \bm{h}_k) \\
&~~~~~(1 \leq i \leq n; 1 \leq j \leq k \leq n)
\end{split}
\end{equation}
Specifically, for word-based SRL, we only need to set the length of candidate arguments to be 1.

\subsection{MLP Scorer}
We employ the MLP scorers as the scoring functions to determine whether the candidate predicates or arguments need to be pruned.
Another MLP scorer is employed to compute the score of whether the focused candidate predicate and argument can compose a semantic relation.
\begin{equation}
\begin{split}
\phi_{p}(p) &= \mathbf{w^{\top}_p}\mathbf{{MLP}_p}({\bm{rep}^{p}_i}) \\
\phi_{a}(a) &= \mathbf{w^{\top}_a}\mathbf{{MLP}_a}({\bm{rep}^{a}_{j, k}}) \\
\phi_{r}(p, a) &= \mathbf{w^{\top}_{r}}\mathbf{{MLP}_r}({[\bm{rep}^{p}_i ; \bm{rep}^{a}_{j, k}]})
\end{split}
\end{equation}

\section{Proposed Framework}
% \begin{figure}[!t]
% 	\centering
% 	\includegraphics[scale=0.8]{biaffine-crop.pdf}
% 	\caption{The revision of biaffine parser module in our proposed framework, where the red circles are extracted syntactic representations $\bm{rep}^{syn}_i$.}
% 	\label{fig:biaffine}
% \end{figure}
Our framework includes two modules, a basic SRL module and a dependency parser module, as shown in Figure \ref{fig:srl-framework}.
% Given an input sentence $s$, the framework first pass the input vector of $s$ into the dependency parser and extract syntactic representations, as depicted by the red circles in the dependency parser module.
% Then we concatenate the syntactic representations with the original input of the SRL module, as shown by the red lines and circles in the right part of Figure \ref{fig:srl-framework}.
In this section, we will first describe the architecture of the employed dependency parser, and then illustrate the integration of the syntactic parser into the basic SRL model.

\subsection{Dependency Parser Module}
We employ the state-of-the-art biaffine parser proposed by \citet{dozat2016deep} as the dependency parser module in our framework, 
as shown by the left part of Figure \ref{fig:srl-framework}.
In order to better fit the dependency parser into our framework, we make some modifications on the original model architecture.
First, we remove the Part-of-Speech (PoS) tagging embeddings and add the Chinese character CNN representations, so the resulting input representation is the same as the SRL module.
Second, we substitute the BiLSTMs in the original biaffine parser with the same 3-layer highway BiLSTMs used in our SRL module.
% The remaining components of biaffine parser remain unchanged.
The biaffine scorer is proposed to compute the score of candidate syntactic head and modifier, which remains unchanged.

\begin{figure}[!t]
	\centering
	\includegraphics[scale=1.0]{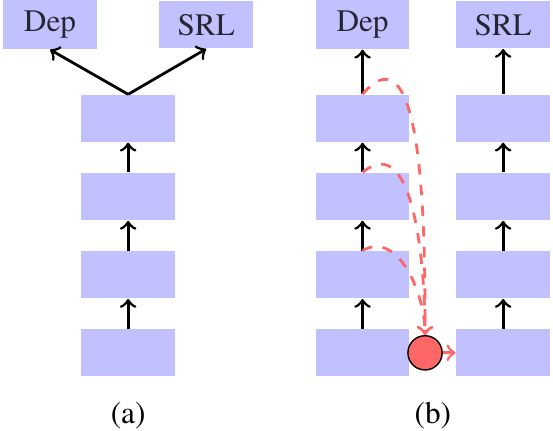}
	\caption{Comparison between the architectures of hard parameter sharing (a) and our proposed implicit representation integration (b).}
	\label{fig:mtl-com}
\end{figure}
\subsection{Details of Integration}\label{sec_hps}
Multi-task learning (MTL) approaches can effectively exploit the standard dependency trees to improve the SRL performance, regarding dependency parsing as the auxiliary task.
% It's a straightforward idea to employ the multi-task learning (MTL) approaches to boost the SRL performance with the dependency parsing as a auxiliary task.
Hard parameter sharing \cite{ruder2017overview} is the most commonly used method in MTL, which shares several common layers between tasks and keeps the task-specific output layers, as illustrated by the sub-figure a in Figure \ref{fig:mtl-com}.
% Specifically, most previous works in the neural network models try to share the input layer and BiLSTM encoder in their frameworks.
We propose a better way to integrate the focused two tasks in this work.
In the following, we will describe the intuition and details of the integration on the two tasks.

As is well known, the hard parameter sharing approach can provide representations for all the shared tasks and reduce the probability of overfitting on the main task.
However, this kind of sharing strategy somewhat weakens the representation framework maintains distinct model parameters for each task, due to the neutralization of knowledge introduced by the auxiliary task.
% So, we present a better way, which can provide the task-specific representation and integrate the information of auxiliary tasks.
Different from the hard parameter sharing strategy, we propose to integrate the syntactic information into the input layer of the basic SRL module, as illustrated by sub-figure b of Figure \ref{fig:mtl-com}.
And Figure \ref{fig:srl-framework} shows the detailed architecture.
First, we extract all the 3 BiLSTM hidden outputs of the dependency parser as the syntactic representations.
Second, we employ the normalized weights to sum the extracted representations as the final syntactic representation for word $w_i$, denoted as $\bm{rep}^{syn}_i$.
Formally, 
\begin{equation}
\bm{rep}^{syn}_i = \sum_{1 \leq j \leq N} \alpha_j * \bm{h}^j_i
\end{equation}
where N is the layer number of the highway BiLSTMs, and $\alpha_j$ is the $j-$th softmax weight.
Finally, the extracted syntactic representations are fed into the input layer of the SRL module, and concatenated with the original SRL module input.
We design this framework for several considerations:
1) the proposed framework keeps the own model parameters for each task, thereby maximizing task-specific information for the main task,
2) the dependency parser module can be updated by the gradients returned from the extracted syntactic representations, which can encourage it to produce semantic preferred representations.

\subsection{Training Objective}
Given the sets of SRL data $\mathcal{S}$ and dependency data $\mathcal{D}$, the framework loss function is defined as the sum of the negative log-likelihood loss of the two tasks:
\begin{equation}
\begin{split}
-\Big (&\sum_{{(Y^*_s, X_s)} \in \mathcal{S}}\log P(Y^*_s|X_s) \\
 &+ %\frac{|\mathcal{S}|}{|\mathcal{D}|}
 \alpha \sum_{(Y^*_d, X_d) \in \mathcal{D}}\log P(Y^*_d|X_d) \Big )
\end{split}
\end{equation}
where $Y^*_s$ and $Y^*_d$ are gold semantic and syntactic structures respectively, and $\alpha$ is a corpus weighting factor to control the loss contribution of the dependency data in each batch as discussed in the experiments. 
%We randomly select the same amount of dependency data as the focused SRL data for each iteration.

\section{Experiments}
\subsection{Settings}
We evaluate the proposed MTL framework on two commonly used benchmark datasets of Chinese: Chinese Proposition Bank 1.0 (CPB1.0) (span-based) \cite{xue2008labeling} and CoNLL-2009 (word-based) \cite{hajivc2009conll}.
Following previous works, we report the results of span-based SRL in two setups: \textit{pre-identified predicates} and \textit{end-to-end}. %since the basic SRL model can jointly predict the predicates and their arguments.
For word-based SRL, we only report the results in the  \textit{pre-identified predicates} setting. 
Following \citet{roth2016neural}, we employ the mate-tools\footnote{\scriptsize{\url{https://code.google.com/archive/p/mate-tools/}}} \cite{bjorkelund2010high} for the predicate disambiguation, which achieves 94.87\% and 94.91\% F1 scores on the CoNLL-2009 Chinese development and test data respectively.

\textbf{Dependency Parsing Data.}
% Most previous syntax-aware methods \cite{roth2016neural, he2018syntax, xia2019syntax} mainly exploit the automatic syntactic trees produced by a pre-trained syntactic parser, which usually have limited syntactic information and systematic errors.
% Moreover, the quality of these syntactic trees is heavily influenced by the performance of the pre-trained parser.
We employ the Chinese Open Dependency Treebank\footnote{\scriptsize{\url{http://hlt.suda.edu.cn/index.php/CODT}}} constructed at Soochow University. The treebank construction project aims to continually build a large-scale Chinese dependency treebank that covers up-to-date texts from different domains and sources \cite{codt-nlpcc19}.
So far, CODT contains 67,679 sentences from 9 different domains or sources.
%, i.e., HIT-CDT (10,xxx), .... [明月提供一下]
% 9个领域，BC包含两个来源，HIT-CDT和Penn-CTB
%BC-train: 16339; comment-train: 6885; content-train: 5129; dialog-train: 6897; finance-train: 8738; PKU-train: 10760; renren-train: 11286; ZX-train: 1645

\textbf{BERT Representations.} 
Recently, BERT (Bidirectional Encoder Representations from Transformers) is proposed by \citet{devlin2018bert}, which makes use of Transformers to learn contextual representations between words.
In this paper, we use the pre-trained Chinese model\footnote{\scriptsize{\url{https://github.com/google-research/bert\#pre-trained-models}}} to extract the BERT representations for our span-based and word-based SRL datasets.
We extract the fixed BERT representations from the last four hidden layers of the pre-trained model.
Finally, we also employ the normalized weighted sum operation to obtain the final BERT representation for each word $w_i$, denoted as $\bm{rep}^{BERT}_i$.

\textbf{Hyperparameters.} 
% For the parameter settings in the SRL module, we mostly follow the work of \citet{he2018jointly}.
We employ word2vec \cite{mikolov2013distributed} to train the Chinese word embeddings on the Chinese Gigaword dataset\footnote{\scriptsize{\url{https://catalog.ldc.upenn.edu/LDC2003T09}}}.
The Chinese char embeddings are randomly initialized, and the dimension is 100.
We employ the CNN to get the Chinese char representations, which has window size of 3, 4 and 5, and the output channel size is 100.
% Then, we use 3 layer of stacked highway BiLSTMs with 200 dimension hidden size.
% Over the highway BiLSTMs, we employ the mean operation to get the representation of candidate arguments.
% As for the candidate predicate of word $w_i$, we directly exploit the output of BiLSTMs $h_i$ as its representation.
% With the representations of predicates and arguments, two MLP layers with ReLU activation functions are employed to score the candidate predicates and arguments.
For other parameter settings in the SRL module, we mostly follow the work of \citet{he2018jointly}.
As for the pruning of candidate predicates and arguments, we choose the pruning ratios according to the training data, using $\lambda_p = 0.4$ for predicates and $\lambda_a = 0.8$ for arguments with up to 30 words.
% Finally, we use a MLP scorer to compute the probability of the predicted predicates and arguments over the semantic label space $\mathcal{R}$.
% For the biaffine dependency parser module, we employ the same model input as the SRL module, so as to keep the consistency of the framework.
% In addition, we also employ the same 3 layer of stacked highway BiLSTMs to encode the input representations, which is different with original biaffine parser.
% The reset of biaffine parser module remains consistent with \citet{dozat2016deep}.

\textbf{Training Criterion.}
We choose Adam \cite{kingma2014adam} optimizer with 0.001 as the initial learning rate and 0.1\% as the decay rate for every 100 steps. 
%We separately feed SRL and dependency batches in turn. For example, we use a SRL batch at the $i$-th step and then use a dependency batch at the $i+1$-th step.
%We randomly shuffle the SRL and dependency samples and then batch them with at most 40 samples or 700 words.
Each data batch is composed of both SRL and dependency instances. 
We randomly shuffle the SRL and dependency training datasets if the smaller SRL data is used up. 
%The framework trains the SRL and dependency batches in turn. %updating the gradients separately when each batch is computed.
All baseline models are trained for at most 180,000 steps, and 100,000 steps for other models.
In addition, we pick the best model on the development data for testing.
We apply 0.5 dropout to the word embeddings and Chinese character representations and 0.2 dropout to all hidden layers.
We employ variational dropout masks that are shared across all timesteps \cite{gal2016theoretically} for the highway BiLSTMs, with 0.4 dropout rate.

\textbf{Evaluation.} 
We adopt the official scripts provided by CoNLL-2005\footnote{\scriptsize{\url{http://www.cs.upc.edu/~srlconll/st05/st05.html}}} and CoNLL-2009\footnote{\scriptsize{\url{https://ufal.mff.cuni.cz/conll2009-st/scorer.html}}} for span-based and word-based SRL evaluation, respectively.
We conduct significant tests using the Dan Bikel's randomized parsing evaluation comparer.

\subsection{Syntax-aware Methods}
To illustrate the effectiveness and advantage of our proposed framework\footnote{\scriptsize{\url{https://github.com/KiroSummer/A_Syntax-aware_MTL_Framework_for_Chinese_SRL}}} (Integration of Implicit Representations, IIR), we conduct several experiments with the recently employed syntax-aware methods on CPB1.0 dataset for comparison:

\begin{itemize}
    \item \textbf{Tree-GRU} \citet{xia2019syntax} investigate several syntax-aware methods for the English span-based SRL, showing the effectiveness of introducing syntactic knowledge into the SRL task. 
    We only compare with the Tree-GRU method, since the other methods are all predicate-specific and hence not fit into our basic SRL model.
    \item \textbf{FIR} Following \citet{yu2018transition} and \citet{zhang2019syntax}, we extract the outputs of BiLSTMs as the fixed implicit representations (FIR) from a pre-trained biaffine parser.
    In detail, we train the biaffine parser with the same training data used in our framework, and employ the combination of development data of CDT (997 sentences) and PCTB7 (998 sentences) as the development data.
    The biaffine parser achieves 79.71\% UAS and 74.74\% LAS on the combined development data.
    \item \textbf{HPS} We employ the commonly used hard parameter sharing (HPS) strategy of MTL as a strong baseline, which shares the word and char embeddings and 3-layer BiLSTMs between the dependency parser and the basic SRL module.
\end{itemize}

\subsection{Main Results}

\begin{table*}[!t]
	\addtolength{\tabcolsep}{+1.0mm}
	\begin{center}
			\begin{tabular}{l ccc ccc}
			        \toprule
			     %   \hline
			        & \multicolumn{3}{c}{Dev} & \multicolumn{3}{c}{Test} \\ 
					\cmidrule(lr){2-4}		
					\cmidrule(lr){5-7}  
					Methods     &P &R &F1        &P &R &F1 \\ \hline
                    \hline
				    Baseline &81.52 &82.17 &81.85  &80.95 &80.01 &80.48 \\
				    Baseline + Dep (Tree-GRU) &82.35 &80.24 &81.28 &82.10 &78.11 &80.06 \\
				    Baseline + Dep (FIR) &\bf83.56 &83.05 &83.30 &83.38 &81.93 &82.65 \\
				    Baseline + Dep (HPS) &82.58 &\bf84.15 &83.36 &83.22 &\bf83.81 &83.51 \\
				    Baseline + Dep (IIR) &83.12 &83.66 &\bf83.39 &\bf84.49 &83.34 &\bf{83.91} \\
				    % \hline
				    \bottomrule
			\end{tabular}
			\caption{Experimental results of syntax-aware methods we compare on CPB1.0 dataset.}
			\label{table:syntactic-ideas}
	\end{center}
\end{table*}

\textbf{Results of Syntax-aware Methods.}
Table \ref{table:syntactic-ideas} shows the results of these syntax-aware methods on CPB1.0 dataset.
First, the first line shows the results of our baseline model, which only employs the word embeddings and char representations as the inputs of the basic SRL model.
Second, the Tree-GRU method only achieves 80.06 F1 score on the test data, which even didn't catch up with the baseline model.
We think this is caused by the relatively low accuracy in Chinese dependency parsing.
Third, the FIR approach outperforms the baseline by 2.17 F1 score on the test data, demonstrating the effectiveness of introducing fixed implicit syntactic representations.
Forth, the HPS strategy achieves more significant performance by 83.51 F1 score.
Finally, our proposed framework achieves the best performance of 83.91 F1 score among these methods, outperforming the baseline by 3.43 F1 score.
All the improvements are statistically significant ($p < 0.0001$).
From these experimental results, we can conclude that: 1) the quality of syntax has a crucial impact on the methods which depend on the systematic dependency trees, like Tree-GRU, 2) the implicit syntactic features have the potential to improve the down-stream NLP tasks, and 3) learning the syntactic features with the main task performs better than extract them from a fixed dependency parser.

\textbf{Results on CPB1.0.}
\begin{table}[tb]
% 	\renewcommand{\arraystretch}{1.2}
% 	\addtolength{\tabcolsep}{0pt}
	\begin{center}
	       % \begin{small}
			\begin{tabular}{l c}
			        \hline
					\textbf{Methods}      &\textbf{F1} \\
					\hline
					\hline
					\bf Previous Works & \\
					\citet{sun2009chinese} & 74.12 \\
				    \citet{wang2015chinese} & 77.59 \\
				    \citet{sha2016capturing} & 77.69 \\
				    \citet{xia2017progressive} & 79.67 \\
				    \hline
                    \hline
                    \bf Ours &  \\
				    Baseline & 80.48 \\
				    Baseline + Dep (HPS) & 83.51 \\
				    Baseline + Dep (IIR) & \bf{83.91} \\
				    \hline
				    Baseline + BERT   & 86.62 \\
				    Baseline + BERT + Dep (HPS) & 87.03 \\
				    Baseline + BERT + Dep (IIR) & \bf 87.54 \\
				    \hline
			\end{tabular}
% 			\end{small}
				\caption{Results and comparison with previous works on CPB1.0 test set. } 
			\label{table:cpb-compare}
	\end{center}
\end{table} 
\begin{table}[tb]
% 	\addtolength{\tabcolsep}{-1.0mm}
	\begin{center}
	       % \begin{small}
			\begin{tabular}{l cc}
			        \hline
			         &\textbf{Dev} &\textbf{Test} \\
					\cmidrule(lr){2-2}
					\cmidrule(lr){3-3} 
					\textbf{Methods} &\textbf{F1} &\textbf{F1} \\
			        \hline
                    \hline
                    \bf Ours  & & \\
				    Baseline &80.37 &79.29 \\
				    Baseline + Dep (IIR) &\bf82.39 &\bf{81.73}\\
				    \hline
				    Baseline + BERT   &85.30 &85.26 \\
				    Baseline + BERT + Dep (IIR) &\bf85.92 &\bf 85.57 \\
				    \hline
			\end{tabular}
% 			\end{small}
				\caption{F1 scores of end-to-end settings on CPB1.0 test set.} 
			\label{table:cpb-e2e}
	\end{center}
\end{table} 
Table \ref{table:cpb-compare} shows the results of our baseline model and proposed framework using external dependency trees on CPB1.0, as well as the corresponding results when adding BERT representations.
It is clear that adding dependency trees into the baseline SRL model can effectively improve the performance ($p < 0.0001$), no matter whether employ the BERT representations or not.
Especially, our proposed framework (IIR) consistently outperforms the hard parameter sharing strategy.
So we only report the results of our proposed framework in later experiments.
Our final results outperforms the best previous model \cite{xia2017progressive} by 7.87 and 4.24 F1 scores with BERT representations or not, respectively.

Table \ref{table:cpb-e2e} shows the results of our framework in the \textit{end-to-end} setting.
To our best knowledge, we are the first to present the results of \textit{end-to-end} on the CPB1.0 dataset.
We achieve the result of 85.57 in F1 score, which is a strong baseline for later works.
It is clear that our framework can still achieve better results compared with the strong baseline, which employs BERT representations as the external input.

\begin{table}[!tb]
	\addtolength{\tabcolsep}{-1.0mm}
	\begin{center}
	       % \begin{small}
			\begin{tabular}{l ccc}
			        \hline
					\textbf{Methods}   & \textbf{P} & \textbf{R} &\textbf{F1} \\
					\hline
					\hline
					\bf Previous Works & & & \\
					\citet{roth2016neural} & 83.2 &75.9 &79.4 \\
					\citet{marcheggiani2017simple}  &84.6 &80.4 &82.5 \\
					\citet{he2018syntax} &84.2 &81.5 &82.8 \\
					\citet{cai2018full} &84.7 &84.0 & 84.3 \\
				    \hline
                    \hline
                    \bf Ours &  & & \\
                    Baseline  &83.7 &84.8 &84.2 \\
                    Baseline + Dep (IIR) &\bf84.6 &\bf85.7 &\bf85.1 \\
                    \hline
                    Baseline + BERT &87.8 &\bf89.2 &\bf88.5 \\
                    Baseline + BERT + Dep (IIR) &\bf88.0 &89.1 &\bf88.5 \\
                    %Baseline+BERT+DepA(SWS) &87.8 &89.5 &88.6 \\
				    \hline
			\end{tabular}
% 			\end{small}
				\caption{Results and comparison with previous works on CoNLL-2009 Chinese test set.} 
			\label{table:conll09-compare}
	\end{center}
\end{table} 
\textbf{Results on CoNLL-2009.}
Table \ref{table:conll09-compare} shows the results of our framework and comparison with previous works on the CoNLL-2009 Chinese test data.
Our baseline achieves nearly the same performance with \citet{cai2018full}, which is an end-to-end neural model that consists of BiLSTM encoder and biaffine scorer.
% In addition, \citet{cai2018full} find that the recently proposed syntax-aware argument pruning algorithm \cite{he2018syntax} has no improvement in their model, which brings a doubt whether syntax still helps when the basic SRL model is rather strong.
Our proposed framework outperforms the best reported result \cite{cai2018full} by 0.8 F1 score and brings a significant improvement ($p < 0.0001$) of 0.9 F1 score over our baseline model.
Our experimental result boosts to 88.5 F1 score when the framework is enhanced with BERT representations.
However, compared with the results in the settings without BERT, the improvement is fairly small (88.53 - 88.47 = 0.06 F1 score, $p > 0.1$)\footnote{\scriptsize{Following previous works, we only retrain the experimental results with one decimal point}} of the proposed framework, which we will discuss in Section \ref{dis-bert}.

\section{Analysis}
In this section, we conduct detailed analysis to understand the improvements introduced by our proposed framework.
% Our proposed framework differs from most previous syntax-aware methods and hard parameter sharing strategy in MTL community.
% To better understand the framework, we conduct two analysis to find where does syntax helps and the impact of syntax datasize.
\subsection{Long-distance Dependencies}\label{discuss-conll09}
% \begin{figure}[!t]
% 	\centering
% 	\includegraphics[scale=0.7]{distance-analysis-cpb-test-crop.pdf}
% 	\caption{F1 regarding to the number of words of sentences on the CPB1.0 test data.}
% 	\label{fig:distance-analysis-cpb}
% \end{figure}
% \begin{figure}[!t]
% 	\centering
% 	\includegraphics[scale=0.7]{distance-conll09-ana-crop.pdf}
% 	\caption{F1 scores regarding to the sentence length of the CoNLL-2009 Chinese test data.}
% 	\label{fig:distance-analysis-conll09}
% \end{figure}
\begin{figure}[!tb]
	\centering
	\includegraphics[scale=0.7]{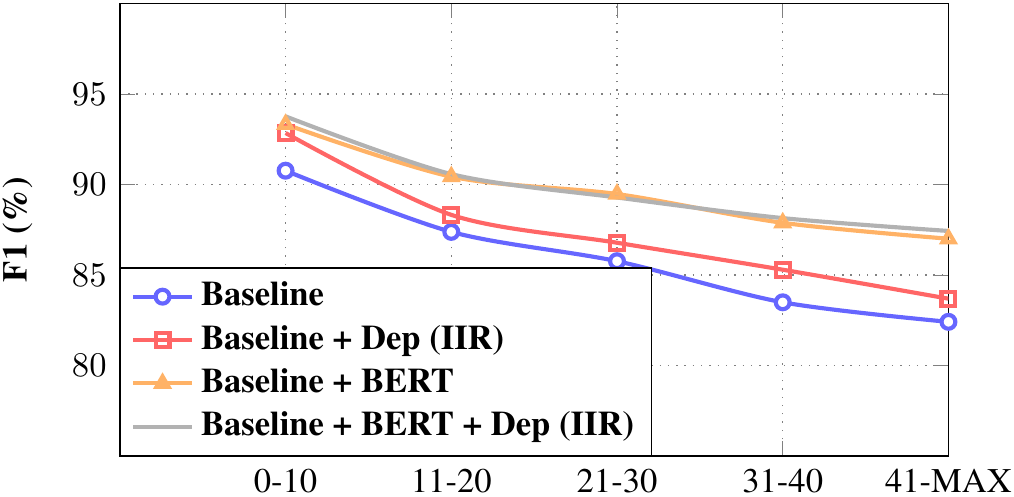}
	\caption{F1 scores regarding to the sentence length of the CoNLL-2009 Chinese dev data.}
	\label{fig:distance-analysis-conll09}
\end{figure}
To analyze the effect of the proposed framework regarding to the distance of sentence lengths, we report the F1 scores of different sets of sentence lengths, as shown in Figure \ref{fig:distance-analysis-conll09}.
We can see that improvements are obtained for nearly all sets of sentences,
especially on the sentences with long-distance.
It demonstrate that \emph{syntactic knowledge is beneficial for SRL and effective to capture long-distance dependencies}.

\subsection{Improvements on Semantic Roles}
\begin{figure}[!tb]
	\centering
	\includegraphics[scale=0.6]{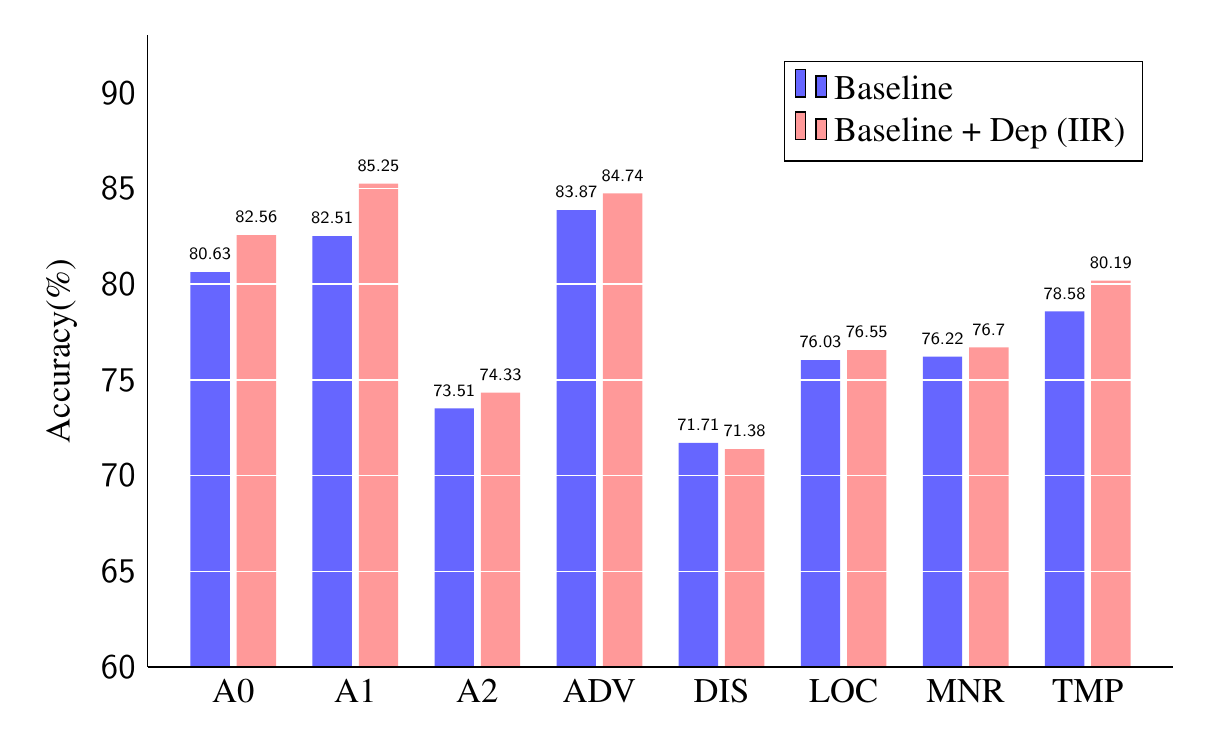}
	\caption{Accuracy comparison of different semantic roles between \textit{Baseline} and \textit{Baseline + Dep (IIR)} on CoNLL-2009 dev data.}
	\label{fig:f1-semantic-roles}
\end{figure}

To find which semantic roles benefit from our syntax-aware framework, we report the F1 scores on several semantic role labels in Figure \ref{fig:f1-semantic-roles}.
We can see that syntax helps most on the \textit{A0} and \textit{A1} roles, which is consistent with the intuition that the semantic \textit{A0} and \textit{A1} roles are usually the syntactic subject and object of a verb predicate.
Other adjunct semantic roles like \textit{ADV}, \textit{LOC}, \textit{MNR} and \textit{TMP} all benefit from the introduction of syntactic information.
There is an interesting finding that the \textit{DIS} role obtains worse performance when introduce syntactic information.
We conduct error analysis on this phenomena, and we found that the framework mostly confuses \textit{DIS} with \textit{ADV}.
The possible reason is that the two semantic roles are both labeled as \textit{adv} in syntax.

\begin{figure}[!tb]
	\centering
	\includegraphics[scale=0.6]{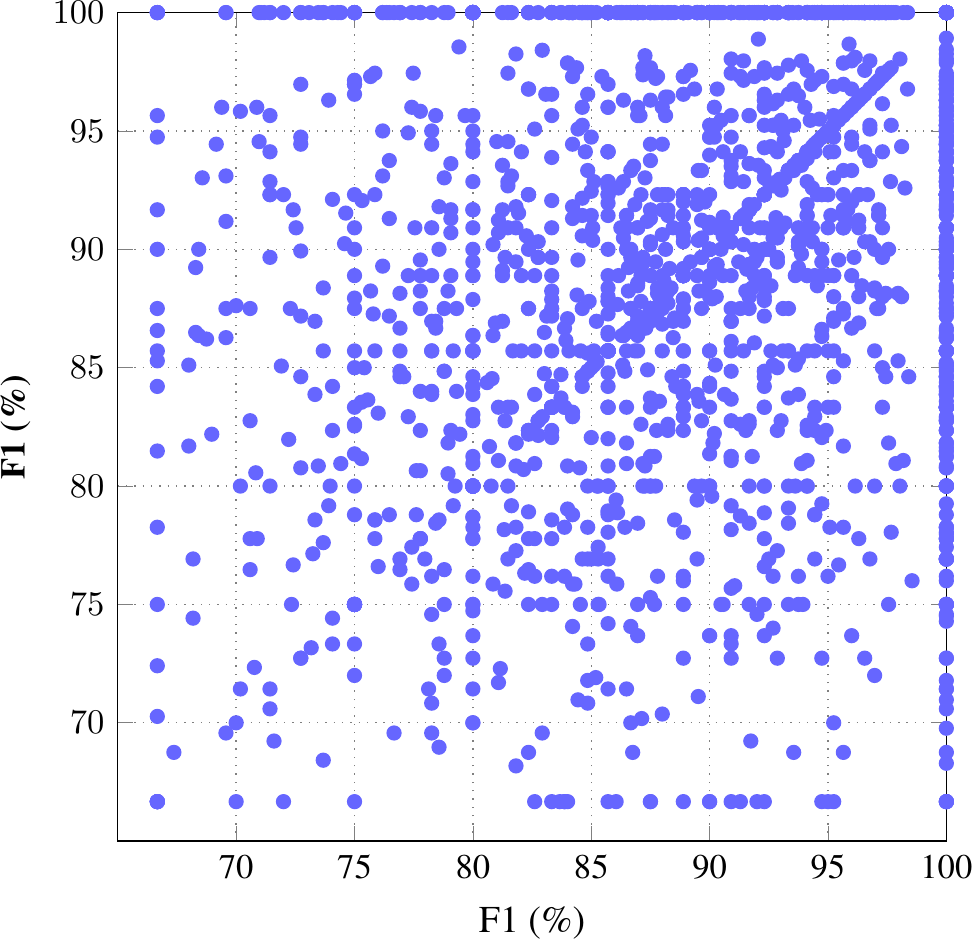}
	\caption{Sentence F1 scores comparison on CoNLL-2009 Chinese test data, where the x axis presents the F1 scores of \textit{Baseline + BERT} and y axis shows the F1 scores of \textit{Baseline + BERT + Dep (IIR)}, respectively.}
	\label{fig:coordinate-analysis-conll09}
\end{figure}
\subsection{Integration with BERT}\label{dis-bert}

% To analyze where does syntax helps in our proposed framework, we conduct several analysis on the employed datasets.
% Figure \ref{fig:distance-analysis-cpb} shows the performance of our experimental results regarding to the lengths of sentences on the CPB1.0 test data.
% The conclusion is clear that \emph{syntax can consistently help the sentences with longer-distance}.
% Besides, we can also see an interesting finding that \emph{syntax also brings a substantial improvement on the sentences within 10 words}, even when the baseline is enhanced with BERT.

% The findings are similar on the CoNLL-2009 Chinese benchmark, as shown in Figure \ref{fig:distance-analysis-conll09}.
BERT is employed to boost the performance of our basic SRL model and our proposed framework.
Compared with results in the settings without BERT, the improvements of our framework over the BERT-enhanced baseline are fairly small on CoNLL-2009, as shown by the last two lines in Table \ref{table:conll09-compare}.
To analyze the difference between the two models (\textit{Baseline + BERT} and \textit{Baseline + BERT + Dep (IIR)}), we conduct an analysis on the sentence performance comparison between them, which is inspired by \citet{zhang2016transition}.
As shown in Figure \ref{fig:coordinate-analysis-conll09}, we can see that most of the scatter points are off the diagonal line, demonstrating strong differences between the two models.
Based on this finding, how to better integrate syntactic knowledge and BERT representations becomes an interesting and meaningful question, and we leave it for future work.

% \begin{figure}[!t]
% 	\centering
% 	\includegraphics[scale=0.6]{attachment.pdf}
% 	\caption{Accuracy comparison of different semantic roles between \textit{Baseline} and \textit{Baseline + Dep{SWS}} on CoNLL-2009 test data.}
% 	\label{fig:f1-semantic-roles}
% \end{figure}
% \begin{figure}[!t]
% 	\centering
% 	\includegraphics[scale=0.8]{case-study-crop.pdf}
% 	\caption{An example of the semantic roles \textit{A1} and \textit{A2} with the same syntactic dependency.}
% 	\label{fig:case-study}
% \end{figure}

% \subsection{Impact of Syntax Datasize}
% \begin{figure}[!t]
% 	\centering
% 	\includegraphics[scale=0.7]{performance-regarding-to-num-of-dep-data-crop.pdf}
% 	\caption{F1 scores regarding to the data-size of dependency data on the CPB1.0 test data.}
% 	\label{fig:performance-dep-data}
% \end{figure}
% As is well known, the quality and quantity of data have crucial influence on all kinds of NLP tasks, so does the syntax-aware methods for the main  task.
% Figure \ref{fig:performance-dep-data} shows the F1 scores of our proposed framework regarding to different size of the shuffled employed dependency data.
% We can see that the performance of SRL improves solidly, along with the size of dependency data.
% This demonstrates that exploiting syntactic data is beneficial for SRL ,and it makes sense to construct large-scale syntactic data.

\section{Related Work}
Traditional discrete-feature-based SRL works \cite{swanson2006comparison, zhao2009semantic} mainly make heavy use of syntactic information.
Along with the impressive development of neural-network-based approaches in the NLP community, much attention has been paid to build more powerful neural model without considering any syntactic information.
\citet{zhou2015end} employ deep stacked BiLSTMs and achieve strong performance for span-based English SRL.
\citet{he2017deep} extend their work \cite{zhou2015end} by employing several advanced practices in recent deep learning literature, leading to significant improvements.
\citet{tan2017deep} present a strong self-attention based model, achieving significant improvements.
Inspired by the span-based model proposed by \citet{lee2017end} for coreference resolution, \citet{he2018jointly, ouchi2018span} present similar span-based models for SRL which can exploit span-level features.
For word-based SRL, \citet{marcheggiani2017simple} propose a simple and fast syntax-agnostic model with rich input representations.
\citet{cai2018full} present an end-to-end model with BiLSTMs and biaffine scorer to jointly handle the predicate disambiguation and the argument labeling sub-tasks.

Apart from the above syntax-free works, researchers also pay much attention on improving the neural-based SRL approaches by introducing syntactic knowledge.
\citet{roth2016neural} introduce the dependency path embeddings to the neural-based model and achieve substantial improvements.
\citet{marcheggiani2017encoding} employ the graph convolutional neural networks on top of the BiLSTM encoder to encode syntactic information.
\citet{he2018syntax} propose a k-th order argument pruning algorithm based on systematic dependency trees.
\citet{strubell2018linguistically} propose a self-attention based neural MTL model which incorporate dependency parsing as a auxiliary task for SRL.
\citet{swayamdipta2018syntactic} propose a MTL framework using hard parameter strategy to incorporate constituent parsing loss into semantic tasks, i.e. SRL and coreference resolution, which outperforms their baseline by +0.8 F1 score.
\citet{xia2019syntax} investigate and compare several syntax-aware methods on span-based SRL, showing the effectiveness of integrating syntactic information.

Compared with the large amount of works on English SRL, Chinese SRL works are rare, mainly because of the limitation of datasize and lack of attention of Chinese researchers.
\citet{sun2009chinese} treat the Chinese SRL as a sequence labeling problem and build a SVM-based model by exploiting morphological and syntactic features.
\citet{wang2015chinese} build a basic BiLSTM model and introduce a way to exploit heterogeneous data by sharing word embeddings.
% \citet{sha2016capturing} employ a quadratic optimization method by capturing two kinds of argument relationships to enhance the model of \citet{wang2015chinese}.
\citet{xia2017progressive} propose a progressive model to learn and transfer knowledge from heterogeneous SRL data.
The above works are all focus on the span-based Chinese SRL, and we compare with their results in Table \ref{table:cpb-compare}.
Different from them, we propose a MTL framework to integrate implicit syntactic representations into a simple unified model on both span-based and word-based SRL, achieving substantial improvements.
%Furthermore, our proposed framework also applies to English SRL.

In addition to the hard parameter sharing strategy that we discuss in Section \ref{sec_hps}, partial parameter sharing strategy is also a commonly studied approach in MTL and domain adaptation.
\citet{kim2016frustratingly} introduce simple neural extensions of feature argumentation by employing a global LSTM used across all domains and independent LSTMs used within individual domains.
\citet{peng2017deep} explore a multitask learning approach which shares parameters across formalisms for semantic dependency parsing.
In addition, \citet{peng2018learning} present a multi-task approach for frame-semantic parsing and semantic dependency parsing with latent structured variables.

\section{Conclusion}
This paper proposes a syntax-aware MTL framework to integrate implicit syntactic representations into a simple unified SRL model.
The experimental results show that our proposed framework can effectively improve the basic SRL model, even when the basic model is enhanced with BERT representations.
Especially, our proposed framework is more effective at utilizing syntactic information, compared with the hard parameter sharing strategy of MTL.
By utilizing BERT representations, our framework achieves new state-of-the-art performance on both span-based and word-based Chinese SRL benchmarks, i.e. CPB1.0 and CoNLL-2009 respectively.
Detailed analysis shows that syntax helps most on the long sentences, because of the long-distance dependencies captured by syntax trees.
Moreover, the comparison of sentence performance indicates that there is still a lot of work to do to better integrate syntactic information and BERT representation.

% \section*{Acknowledgments}
\section*{Acknowledgments}
We thank our anonymous reviewers for their helpful comments.
This work was supported by National Natural Science Foundation of China (Grant No. 61525205, 61876116,  61432013) and a project funded by the Priority Academic Program Development of Jiangsu Higher Education Institutions.

\bibliography{emnlp-ijcnlp-2019}
\bibliographystyle{acl_natbib}
\end{document}